\documentclass[fullpaper,cameraready]{nldl}

\renewcommand{\DOI}{10.7557/18.6796}

\usepackage[utf8]{inputenc}
\usepackage{url}
\usepackage{graphicx}
\usepackage{amsmath, amssymb}
\usepackage{mathtools}
\usepackage{authblk}
\usepackage{chemformula}
\usepackage{multirow}
\usepackage{xspace}
\usepackage{makecell}

\usepackage{algorithm}
\usepackage{algorithmic}
\usepackage{booktabs}

\usepackage[square,numbers]{natbib}

\graphicspath{ {./images/} } 
\newcommand\X{\mathbf X}
\newcommand\E{\mathbf E}
\newcommand\M{\mathbf M}
\newcommand\R{\mathbb R}
\newcommand\Lm{\mathcal L}

\newcommand\z{\mathbf z}
\newcommand\g{\mathbf g}

\newcommand{\ie}{\emph{i.e.}\xspace}
\newcommand{\eg}{\emph{e.g.}\xspace}

\usepackage{hyperref}

\title{Explainability in subgraphs-enhanced Graph Neural Networks}
\author[1]{Michele Guerra\thanks{Corresponding Author: michele.guerra@uit.no}}
\author[2]{Indro Spinelli}
\author[2]{Simone Scardapane}
\author[1,3]{Filippo Maria Bianchi}
\affil[1]{UiT the Arctic University of Norway}
\affil[2]{Sapienza University of Rome}
\affil[3]{NORCE Norwegian Research Centre}

\date{\vspace{-5ex}}

\begin{document}
\nldlmaketitle

\begin{abstract} 

    Recently, subgraphs-enhanced Graph Neural Networks (SGNNs) have been introduced to enhance the expressive power of Graph Neural Networks (GNNs), which was proved to be not higher than the 1-dimensional Weisfeiler-Leman isomorphism test. 
    The new paradigm suggests using subgraphs extracted from the input graph to improve the model's expressiveness, but the additional complexity exacerbates an already challenging problem in GNNs: explaining their predictions. In this work, we adapt PGExplainer, one of the most recent explainers for GNNs, to SGNNs. The proposed explainer accounts for the contribution of all the different subgraphs and can produce a meaningful explanation that humans can interpret. The experiments that we performed both on real and synthetic datasets show that our framework is successful in explaining the decision process of an SGNN on graph classification tasks.

\end{abstract}

\section{Introduction}
    
    Recent work on Graph Neural Networks (GNNs) focused on improving their \emph{expressiveness}, which is the ability of the GNN to distinguish between topologically different graphs. 
    Notably, \cite{xu2018powerful} and~\cite{morris2019weisfeiler} proved that GNNs can be expressive as 1-dimensional Weisfeiler-Leman (WL) isomorphism test~\cite{weisfeiler1968reduction}.
    To design more expressive GNNs, some works took inspiration from more powerful, higher-order $k$-WL tests, while subgraphs-enhanced Graph Neural Networks (SGNNs) try to improve the expressive power of GNNs by reasoning on subgraphs.
    In this work, we focus on a recent SGNN framework called Equivariant Subgraph Aggregation Network (ESAN)~\cite{bevilacqua2021equivariant}. 
    The basic intuition in ESAN is that a structural difference between two non-isomorphic graphs that is not apparent when processing the whole graph with a GNN, could instead be detected by looking at their subgraphs. 
    ESAN represents each graph with a set of subgraphs obtained with some predefined fixed policy, \eg, by removing one node or one edge, which is then processed to provide an embedding of the original graph to be used for the downstream task.

    The message-passing framework allows GNNs to learn graph representations that capture both node features and relational information of the graph. 
    Since GNNs perform nonlinear transformations and aggregations, their predictions are not immediately intelligible.
    Therefore, a challenging yet important task is to develop \emph{explainability} tools to fully understand the GNN behaviour, allowing researchers to interpret the decision process in critical applications, to fix mistakes, or to improve the GNN design. 
    Explaining the context of GNNs means highlighting both patterns in node features and node relations that mostly correlate with the output. 
    One of the first attempts was made in~\cite{ying2019gnnexplainer} with GNNExplainer, which produces for each input a subgraph and a subset of node features. 
    Since the output of GNNExplainer is instance-dependent, it struggles to capture the decision process of the GNN for the whole downstream task. The only proposed workaround to explain a set of predictions is to aggregate multiple individual explanations.
    On the other hand, a more recent method called PGExplainer~\cite{luo2020parameterized} can learn how to sample subgraphs that highlight the most relevant parts of the input that influence the GNN output.

    So far, explainers have been applied only to GNNs operating on a single graph, and an interpretability framework for models consuming multiple subgraphs is still missing.
    To fill this gap, in this work we extend the PGExplainer framework to provide a \emph{post-hoc} explainer for SGNN models that, like ESAN, directly operate on subgraphs.
    We tackle two main challenges: i) how to interface PGExplainer to the
    ESAN framework, so that it can handle subgraphs, and ii) how to meaningfully combine the learned subgraph explanations to obtain a single explanation for the original graph. 
    The results show that our framework successfully overcomes the complexity added by the use of subgraphs, and it is able to produce meaningful explanations for an SGNN.

\section{Preliminaries}

\subsection{Notation}
    Using standard notation, a graph is a pair $\mathcal{G}=(\mathcal{V},\mathcal{E})$, where $\mathcal{V}$ is the set of vertices and $\mathcal{E}\subseteq\mathcal{V}\times\mathcal{V}$ is the set of edges. The structure of a graph can be also represented with an adjacency matrix $\mathbf{A}$ where $A_{ij}=1$ if $(i,j)\in\mathcal{E}$, and it's zero otherwise. Associated to each node we have a $d$-dimensional feature vector collected in a matrix $\mathbf{X}$.
    In a graph classification setting, each graph $\mathcal{G}_i$ is associated with a label $y_i$ and the dataset consists of  pairs $\{\mathcal{G}_i, y_i\}$.
    
\subsection{Graph Neural Networks}
    GNNs generalize the approach of convolutional neural networks to the graph domain, using message-passing to aggregate information flowing along edges to learn node embeddings~\cite{kipf2016semi}. The formula of a generic message-passing layer in a GNN is
    \begin{align*}
        \mathbf{X}_i^{(k)}=\gamma^{(k)}\left( \mathbf{X}_i^{(k-1)}, \Box_{j} \phi^{(k)} \left( \mathbf{X}_i^{(k-1)}, \mathbf{X}_j^{(k-1)} \right) \right),
    \end{align*}
    where $k$ is the layer index, $\phi$ is an MLP representing the propagation step that computes the message of each edge in the neighbourhood of node $i$ ($\mathcal{N}(i)$), $\Box$~is the aggregation operator, and $\gamma$ is another MLP which updates the node feature $\X_i^{(k)}$.

    Almost every MP layer can be seen as a generalization of the above expression.
    For example, GIN~\cite{xu2018powerful} implements message-passing as follows:
    \begin{equation}
    \label{eq:gin}
        \mathbf{X}_i^{(k)}=\gamma^{(k)}\!\left(\!(1+\epsilon)\mathbf{X}_i^{(k-1)} +\! \sum_{j\in\mathcal{N}(i)}\!\mathbf{X}_j^{(k-1)}\!\right),
    \end{equation}
    where $\epsilon$ is a learnable parameter.
    
\subsection{ESAN framework}
\label{sec:esan}
    A key element in the ESAN framework is a pre-processing step where a \emph{bag} of $M$ subgraphs $\{\mathcal{S}_1,\dots,\mathcal{S}_M\}$ is extracted from each graph $\mathcal{G}$ in the dataset.
    The subgraphs are obtained with one of the following policies:
    \begin{itemize}
        \item Edge Deletion: one edge is randomly removed;
        \item Node Deletion: one node is randomly removed, along with all its incident edges;
        \item Ego Network: for each node, one builds an ego-network of predefined depth $k$, which is the neighbourhood of order $k$;
        \item Ego Network+: a slightly modified version of the ego-network approach, where the central node is provided with a special feature to distinguish it from the others.
    \end{itemize}

    Afterwards, the bag of subgraphs is used to train a model that consists of three parts: an \emph{Equivariant Feature Encoder} $f_{\text{EFE}}$, a \emph{Subgraph Readout Layer} $f_{\text{SRL}}$, and a \emph{Set Encoder} $f_{\text{SE}}$. The whole ESAN framework can be summarized as follows
    \begin{equation*}
        \mathcal{G} \mapsto \{\mathcal{S}_i\} \xmapsto{f_{\text{EFE}}} \{\E_i\} \xmapsto{f_\text{SRL}} \{\z_i\} \xmapsto{f_\text{SE}} \g.
    \end{equation*}
    The Equivariant Feature Encoder ($f_{\text{EFE}}$) is composed of message-passing layers and is designed to preserve the symmetry with respect to permutations of subgraphs in the bag and of nodes within a graph; in output, it produces for all the nodes in subgraph $\mathcal{S}_i$ an embedding $\E_i \in \R^{N_i \times h}$, where $N_i$ is the number of nodes in $\mathcal{S}_i$ and $h$ is the embedding dimension.
    The node embeddings are then processed by the Readout Layer ($f_\text{SRL}$), which outputs a single feature vector $\z_i$ for each subgraph $\mathcal{S}_i$. 
    Finally, the Set Encoder ($f_\text{SE}$) aggregates all the feature vectors and returns an embedding $\g$ of the original graph $\mathcal{G}$, which is used for the given task (\eg, graph classification).

    The encoder $f_{\text{EFE}}$ comes in two possible main configurations. 
    The most general one is DSS-GNN, where subgraph $\mathcal{S}_i$, with adjacency matrix $\mathbf{A}^{(i)}$ and node features $\mathbf{X}^{(i)}$, is transformed by each layer into
    \[                \E_i=f_{\text{EFE}_1}\!\left(\mathbf{A}^{(i)},\mathbf{X}^{(i)}\right)+f_{\text{EFE}_2}\!\left(\sum_i\,\mathbf{A}^{(i)},\sum_i\,\mathbf{X}^{(i)}\right),
    \]
    where both $f_{\text{EFE}_1}$ and $f_{\text{EFE}_2}$ are a composition of GNN layers and $f_{\text{EFE}_2}$ allows subgraphs in the bag to share information. A simpler variant is DS-GNN, where each subgraph is treated independently so that information is shared only at the end within the Set Encoder. The original paper uses GIN and some small variants to implement the layers in $f_{\text{EFE}_1}$ and $f_{\text{EFE}_2}$.
    
\subsection{PGExplainer}
\label{sec:explainer}
    The intuition behind PGExplainer is that one can split an input graph into two subgraphs, $\mathcal{G}=\mathcal{G}_{ex}+\Delta G$, where $\mathcal{G}_{ex}$ is the explanation subgraph, \ie, it contains most of the information which is relevant for the GNN's prediction. 
    At the core of the explainer, there is an MLP that takes as inputs the graph structure and the node representations produced by the GNN to be explained, and it outputs for each edge a latent variable $\omega_{ij}$. During training, the explainer produces the subgraph $\mathcal{G}_{ex}$ by sampling for each edge $(i,j)\in\mathcal{E}$ a weight $e_{ij}\in(0,1)$ that, when it is zero, masks out the unimportant edges. The authors let $e_{ij}$ follow a Bernoulli distribution, which they reparametrize with a binary concrete distribution~\cite{maddison2017the} to allow gradient computation:
    \begin{equation}
    \label{eq:edge}
        e_{ij} = \sigma((\omega_{ij} + (\log(u)-\log(1-u)))/\tau),
    \end{equation}
    where $\sigma$ is the sigmoid function, $\tau$ is a temperature parameter, and $u\sim\mathcal{U}(0,1)$.

    The loss function of PGExplainer consists of:
    \begin{itemize}
        \item[$\Lm_\text{ce}$] a cross-entropy term between the prediction obtained from the original graph $\mathcal{G}$ and the prediction obtained  from the explanation $\mathcal{G}_{ex}$;
        \item[$\Lm_{\ell_1}$] a $\ell_1$ regularization term which encourages $\mathcal{G}_{ex}$ to be sparse;
        \item[$\Lm_\text{ent}$] an element-wise entropy term, which encourage $e_{ij}$ to be either 0 or 1;
        \item[$\Lm_\text{con}$] an (optional) cross-entropy term between adjacent edges, which encourages the subgraph $\mathcal{G}_{ex}$ to be connected.
    \end{itemize}

    The total loss function is given by
    \begin{equation}
        \label{eq:pgex_loss}
        \Lm_\text{exp} = \Lm_\text{ce} + \Lm_{\ell_1} + \Lm_\text{ent} (+ \Lm_\text{con}).
    \end{equation}

\section{Methods}
    \begin{figure*}[!ht]
        \centering
        \includegraphics[width=0.9\textwidth]{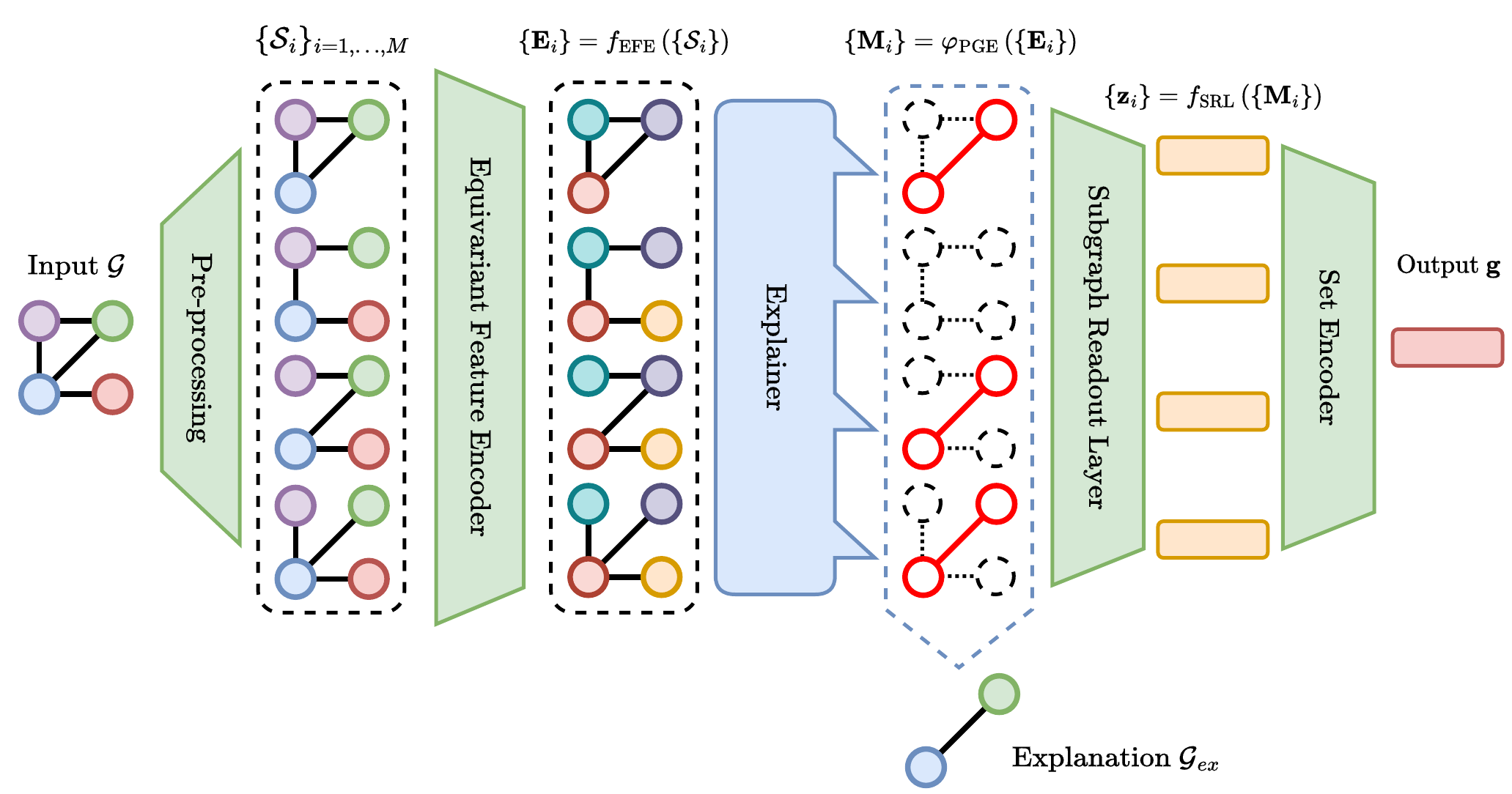}
        \caption{\small Illustration of the workflow to explain ESAN. The green components represent modules from the standard ESAN network, the part in blue is PGExplainer. Starting from the left: the input graph $\mathcal{G}$ is first decomposed in a bag of subgraphs $\{\mathcal{S}_i\}$ according to the chosen policy (in the example, we deleted one edge), then the EFE processes each subgraph providing a representation for all nodes, which PGExplainer uses to sample a bag of masks $\{\M_i\}$. The masked embeddings are then passed to the SRL and SE components, which provide the masked output $\g$ used to train PGExplainer itself. The masks $\M_i$ are then merged to produce the final explanation $\mathcal{G}_{ex}$.}
        \label{fig:scheme}
    \end{figure*}
    A schematic representation of our workflow is pictured in Figure~\ref{fig:scheme}. The source code implementing our workflow is publicly available online.\footnote{\url{https://github.com/MicheleUIT/Explaining_SGNN}}
    
    In the present work we use ESAN in both the configurations described in Sec.~\ref{sec:esan} -- DSS-GNN and DS-GNN -- and we use a modified version of GIN to implement $f_{\text{EFE}_1}$ and $f_{\text{EFE}_2}$.
    The original GIN layer in \eqref{eq:gin} treats the graph edges as binary in computing the message-passing, which makes the whole network agnostic to the edge weights $e_{ij}$ sampled by PGExplainer. 
    To avoid this, we modify the propagation step from $\phi^{(k)}=\mathbf{X}_j^{(k-1)}$ to $\phi^{(k)}=\mathbf{X}_j^{(k-1)}e_{ij}$.
    
    The output of ESAN is an embedding vector representing the whole graph.
    However, as mentioned in Sec.~\ref{sec:explainer}, the explainer does not work with an embedding of the whole input graph but it takes as input the graph structure and the node embeddings. The node embeddings required by the explainer are produced by ESAN at an early stage by the Equivariant Feature Encoder (called $\E_i$ in the figure) before they are passed to the Subgraph Readout Layer that aggregates and makes the node-wise information indistinguishable. 

    For the same reason, the explainer is trained to produce a mask $\M_i$ that explains each individual subgraph $\mathcal{S}_i$. In order to produce an explanation $\mathcal{G}_{ex}$ for the whole input graph, and not just to its subgraphs separately, we have to merge each mask, being careful to keep the correspondence between nodes and edges. The merging could be done by taking, for each edge $(i,j) \in \mathcal{E}$, the average of the $M$ weights $e_{ij}$ learned by the explainer for each replica of $(i,j)$ in each subgraph.
    If the edge $(i,j)$ is missing in one of the subgraphs, we simply set $e_{ij}=0$ for that subgraph.
    Besides the average, the masks of the subgraphs could be merged by taking the maximum value, or by summing them and then rescaling the overall mask so to have values between 0 and 1. 
    We opted for the latter approach because it gives more importance to those edges consistently associated with a high weight $e_{ij}$ across different~masks.

    Our implementation of PGExplainer departs from the original one in some aspects.

    We reformulated the reparametrization trick in~\eqref{eq:edge} as follows
    \[
        e_{ij} = \sigma\left(\frac{\omega_{ij}+\xi}{\tau}\right),
    \]
    with $\xi = -\log(-\log(u))$ and $u\sim\mathcal{U}(0,1)$.
    Empirically, we found that such a reparametrization leads to values in $e_{ij}$ that are slightly closer to 0 or 1.
    
    To circumvent the problem of ``introduced evidence" \cite{dabkowski2017real} generated by the presence of soft masks (\ie, masks with weights lying in the $[0,1]$ interval), we devise a way to force each edge weight during training to assume a binary value in $\{0,1\}$ (commonly called ``hard'' mask) using a straight-through estimator \cite{hinton2012deep} that, at the same time, allows the flowing of gradients. In particular, during forward pass we first sample the soft mask using the reparametrization trick described above; then, we binarize the soft mask by applying a threshold (an operation that is not differentiable) and we obtain a hard mask that is applied to the subgraphs. In the backward step, gradients are computed with respect to the soft mask, ignoring the non-differentiable step that produces hard masks.
    
    We modified the loss function in \eqref{eq:pgex_loss} by dropping the element-wise entropy term $\Lm_\text{ent}$ since, in our experiments, it did not appear to be helpful thanks to the straight-through estimator described above.
    We also noticed that the produced explanations did not require an additional constraint to encourage connectivity and, thus, we did not include regularization terms such as $\Lm_\text{con}$. 
    Therefore, the deployed loss function has only a cross-entropy term between the original classification prediction produced by ESAN and the one given with the masked subgraphs, and the $\ell_1$ regularization term to encourage the sampling of smaller masks:
    \[
    \Lm_\text{exp} = \Lm_\text{ce} + \Lm_{\ell_1}.
    \]

\section{Experiments}

\subsection{Datasets}
    \begin{table}[t]
        \small
         \centering
         \begin{tabular}{c|cc}
            \cmidrule[1.5pt]{1-3}
             Dataset & BA-2motifs & Mutagenicity \\
             Motifs & \raisebox{-.5\height}{\includegraphics[width=0.15\textwidth]{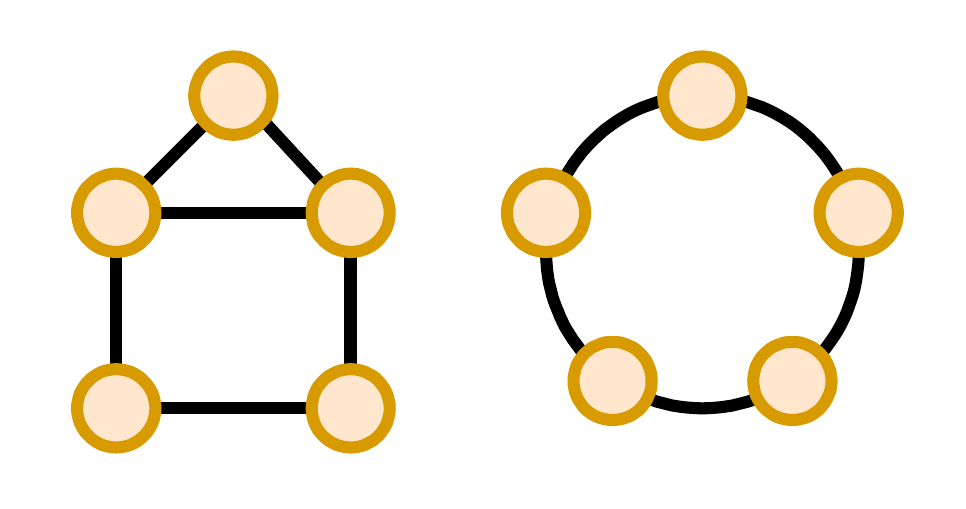}} & \raisebox{-.5\height}{\includegraphics[clip, width=0.15\textwidth]{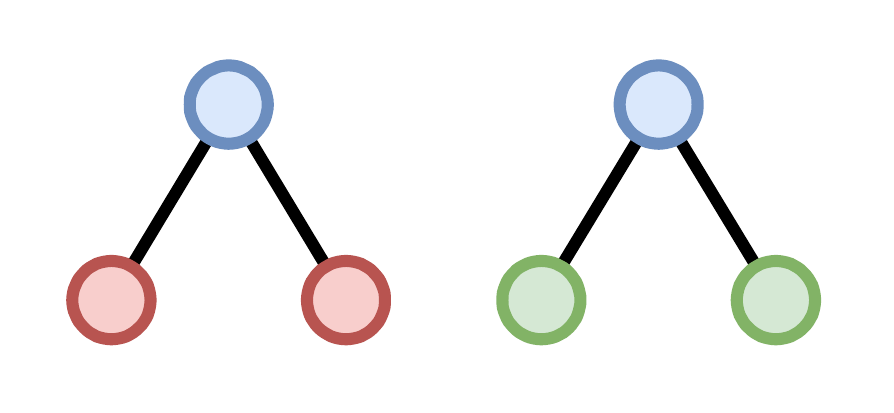}} \\
            \cmidrule[1.5pt]{1-3}
         \end{tabular}
         \caption{\small Description of the two datasets used.}
         \label{tab:datasets}
    \end{table}
    We used two datasets for graph classification that provide a ground-truth explanation for each graph (see the summary in Table~\ref{tab:datasets}):
    \begin{itemize}
        \item a synthetic dataset, BA-2motifs \cite{luo2020parameterized}, which contains $1000$ graphs divided into two classes according to the motif they contain: either a ``house'' or a five-node cycle;
        \item a real-world dataset, Mutagenicity \cite{kazius2005derivation}, which contains $4337$ graphs that represent mutagen drugs, if they have \ch{NH2} or \ch{NO2} groups, and non-mutagen drugs if they don't.
    \end{itemize} 

\subsection{Metrics}
    Finding an explanation motif can be seen as a binary classification problem for edges: either they are part of the explaining motif, or they are not. This justifies the use of AUC as an evaluation of the accuracy of the explainer. In Table~\ref{tab:results} we also report the average size of the explanations, expressed as a percentage with respect to the whole input graph. If it is too large the masked input is likely to produce the same output given by the original graph, but the mask is less explanatory because it contains a lot of uninformative edges. On the other hand, if the size is too small, the masked input could not capture all the relevant information to produce the right output.

\subsection{Results}
    
    \bgroup
    \def\arraystretch{1} 
    \setlength\tabcolsep{.1em} 
    \begin{table}[ht]
        \small
        \centering
        \[
        \begin{array}{c|cccc}
            \toprule
            \text{Dataset} & \text{EFE} & \text{Policy} & \text{Mask size (\%)} & \text{AUC}  \\
            \midrule
            \multirow{8}{*}{\rotatebox[origin=c]{90}{\text{BA-2motifs}}}
                & \text{DS} & \text{ND} & 19\pm6 & 0.88\pm0.14 \\ 
                & \text{DS} & \text{ED} & 33\pm5 & 0.78\pm0.15 \\ 
                & \text{DS} & \text{EN} & 4\pm13 & 0.70\pm0.07 \\ 
                & \text{DS} & \text{EN+} & 17\pm4 & 0.99\pm0.00 \\ 
                & \text{DSS} & \text{ND} & 22\pm1 & 0.99\pm0.00 \\ 
                & \text{DSS} & \text{ED} & 21\pm3 & 0.99\pm0.00 \\ 
                & \text{DSS} & \text{EN} & 1\pm1 & 0.69\pm0.05 \\ 
                & \text{DSS} & \text{EN+} & 0\pm0 & 0.74\pm0.06 \\ 
            \text{Baseline} & \text{--} & \text{--} & 12\pm0 & 0.99\pm0.00 \\
            \midrule          
            \multirow{8}{*}{\rotatebox[origin=c]{90}{\text{Mutagenicity}}}
                & \text{DS} & \text{ND} & 47\pm2 & 0.80\pm0.03 \\ 
                & \text{DS} & \text{ED} & 59\pm8 & 0.77\pm0.04 \\ 
                & \text{DS} & \text{EN} & 6\pm1 & 0.76\pm0.13 \\ 
                & \text{DS} & \text{EN+} & 10\pm0 & 0.88\pm0.01 \\ 
                & \text{DSS} & \text{ND} & 8\pm8 & 0.86\pm0.06 \\ 
                & \text{DSS} & \text{ED} & 68\pm4 & 0.90\pm0.02 \\ 
                & \text{DSS} & \text{EN} & 0\pm0 & 0.58\pm0.03 \\ 
                & \text{DSS} & \text{EN+} & 2\pm2 & 0.60\pm0.06 \\ 
            \text{Baseline} & \text{--} & \text{--} & 30\pm25 & 0.63\pm0.18 \\
            \bottomrule
        \end{array}
        \]
        \caption{\small Results for the two datasets, BA-2motifs and Mutagenicity. DS and DSS represent the two possible configurations of the Equivariant Feature Encoder ($f_\text{EFE}$). The possible policies described in Sec.~\ref{sec:esan} to extract subgraphs are: deleting one node (ND), deleting one edge (ED), ego networks (EN), and ego networks with a distinguishing feature on the node (EN+). Both the AUC and the mask size (as a percentage of edges in the original graph) are reported. Baselines refer to PGExplainer applied to a regular GIN.}
        \label{tab:results}
    \end{table}
    \egroup

    \bgroup
    \setcellgapes{-7pt}
    \makegapedcells
    \begin{table*}[t]
        \small
        \centering
        \begin{tabular}{cc|cccc}
            \toprule
            \multicolumn{2}{c}{Policy} & ND & ED & EN & EN+ \\
            \midrule
            \rotatebox[origin=c]{90}{BA-2motifs}
                & \makecell{\Gape[0.8cm]{\rotatebox[origin=c]{90}{DS}} \\ \Gape[0.8cm]{\rotatebox[origin=c]{90}{DSS}}}
                & \makecell{\includegraphics[width=0.15\textwidth]{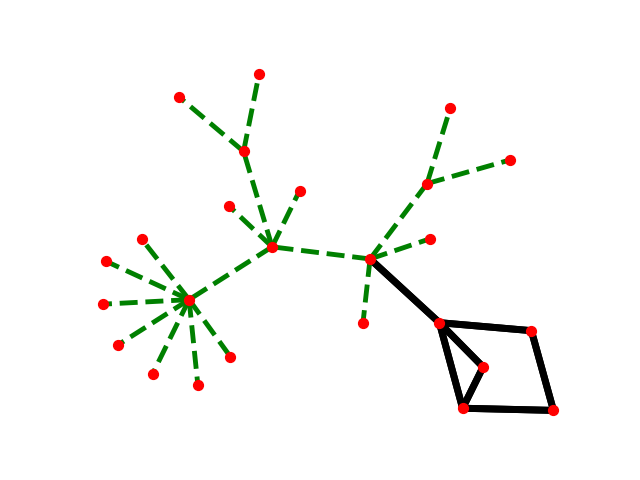} \\ \includegraphics[width=0.15\textwidth]{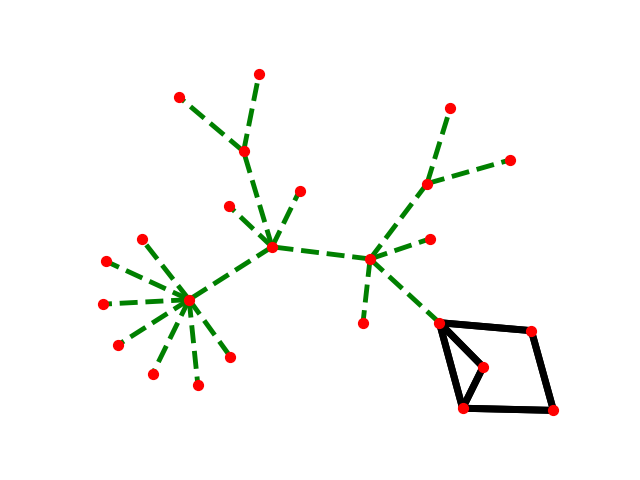}}
                & \makecell{\includegraphics[width=0.15\textwidth]{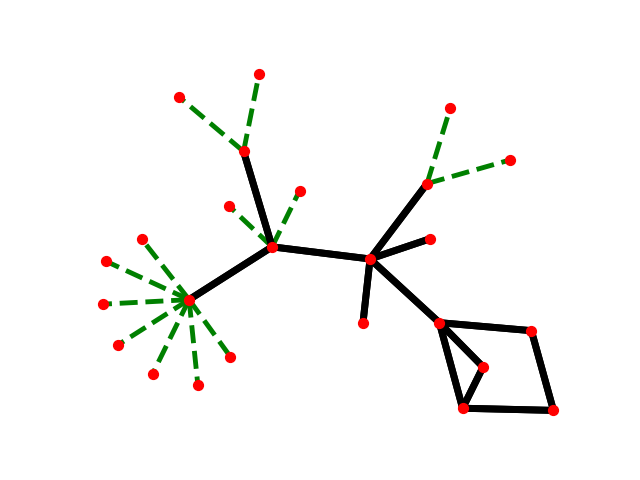} \\ \includegraphics[width=0.15\textwidth]{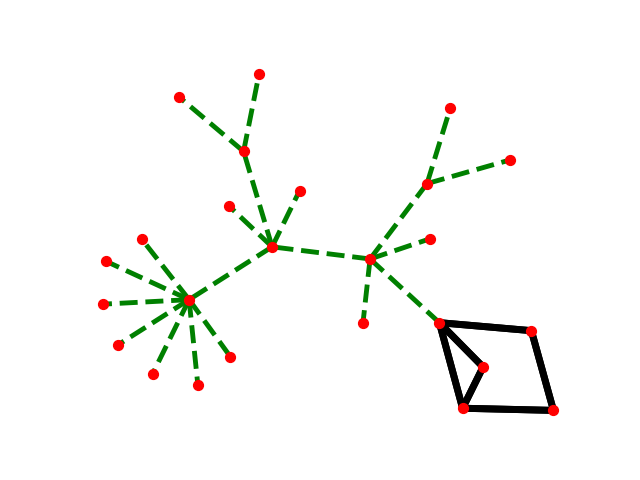}}
                & \makecell{\includegraphics[width=0.15\textwidth]{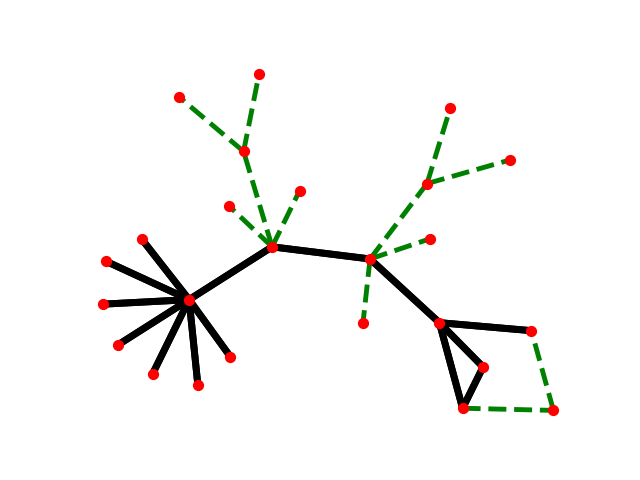} \\ \includegraphics[width=0.15\textwidth]{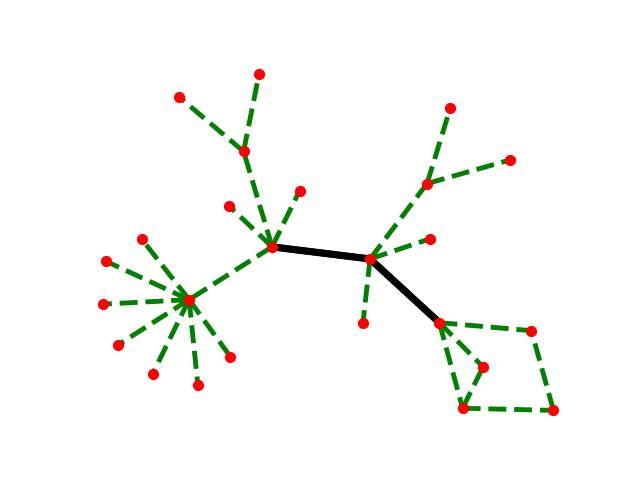}}
                & \makecell{\includegraphics[width=0.15\textwidth]{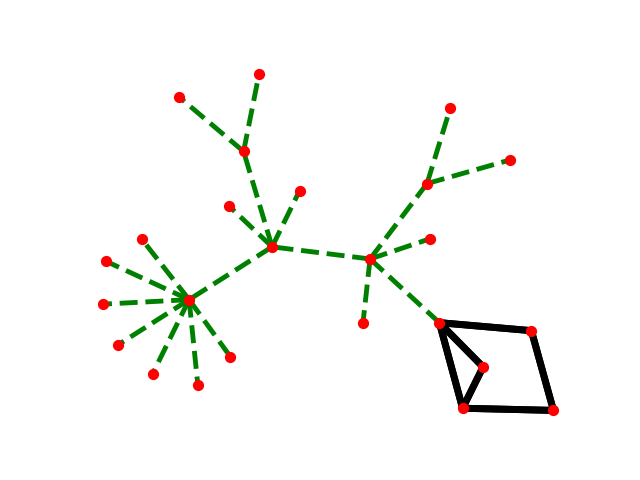} \\ \includegraphics[width=0.15\textwidth]{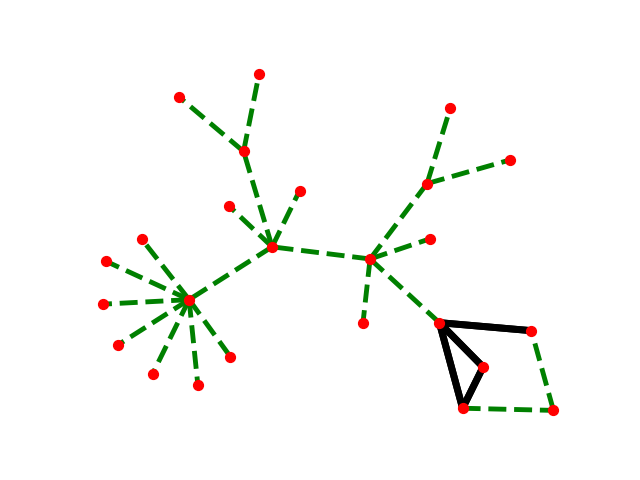}} \\
            \midrule
            \rotatebox[origin=c]{90}{Mutagenicity}
                & \makecell{\Gape[0.8cm]{\rotatebox[origin=c]{90}{DS}} \\ \Gape[0.8cm]{\rotatebox[origin=c]{90}{DSS}}}
                & \makecell{\includegraphics[width=0.15\textwidth]{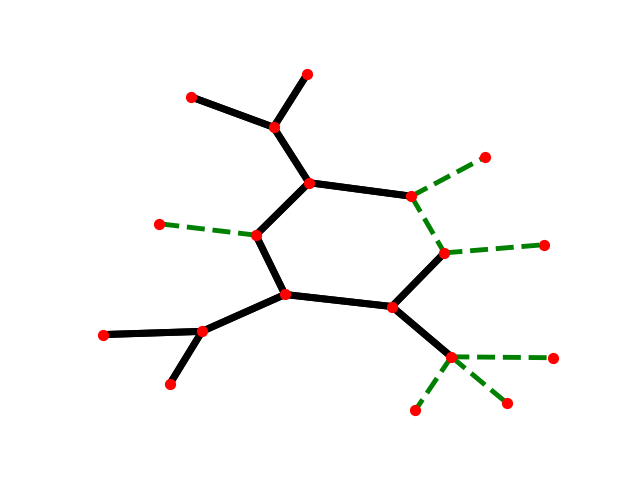} \\ \includegraphics[width=0.15\textwidth]{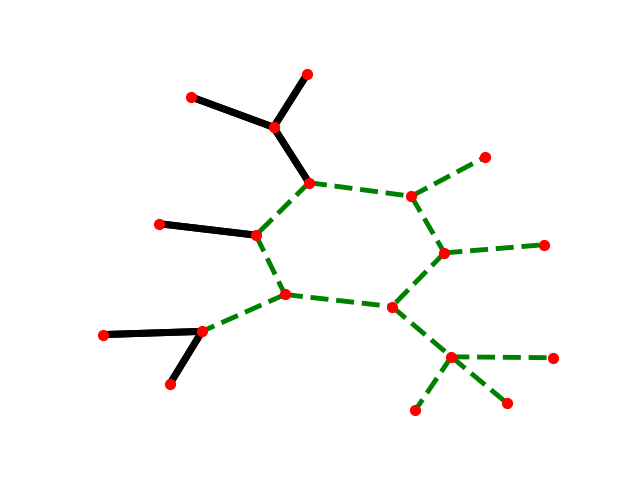}} 
                & \makecell{\includegraphics[width=0.15\textwidth]{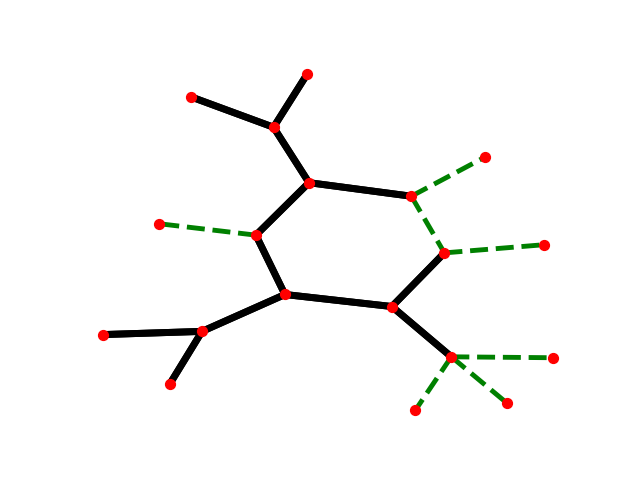} \\ \includegraphics[width=0.15\textwidth]{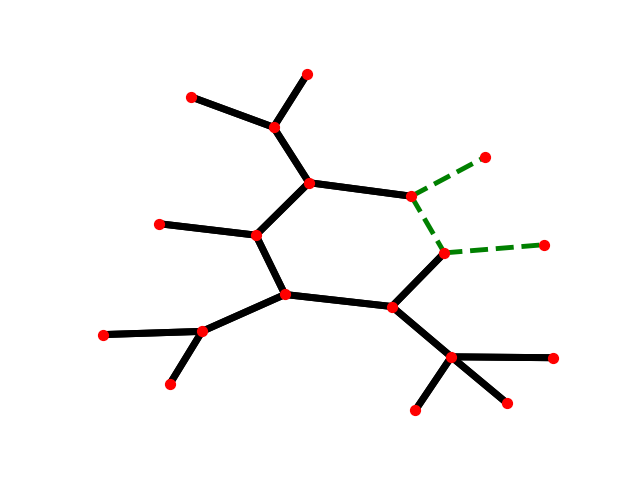}}
                & \makecell{\includegraphics[width=0.15\textwidth]{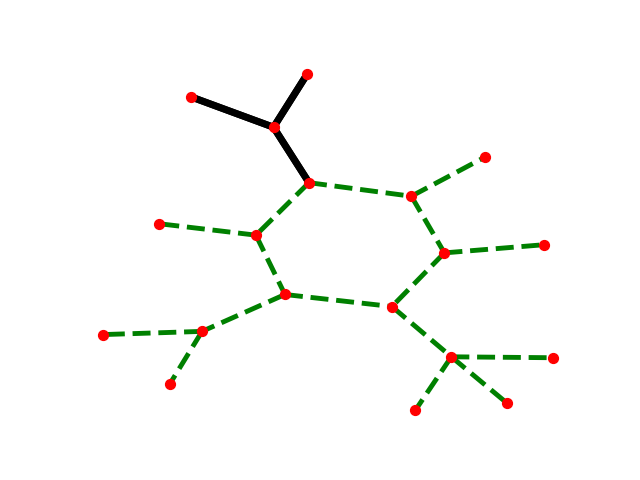} \\ \includegraphics[width=0.15\textwidth]{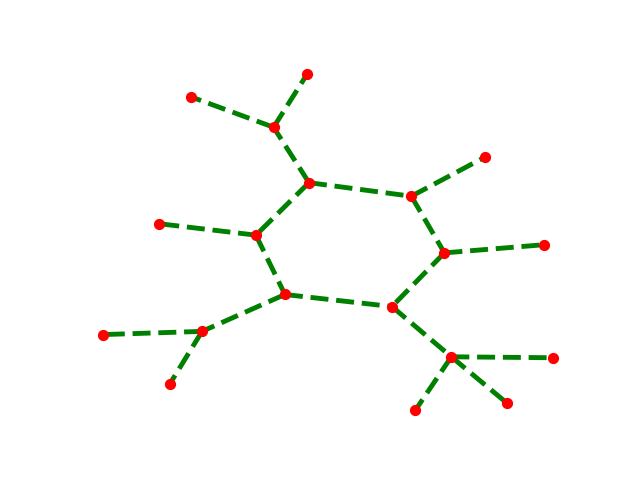}}
                & \makecell{\includegraphics[width=0.15\textwidth]{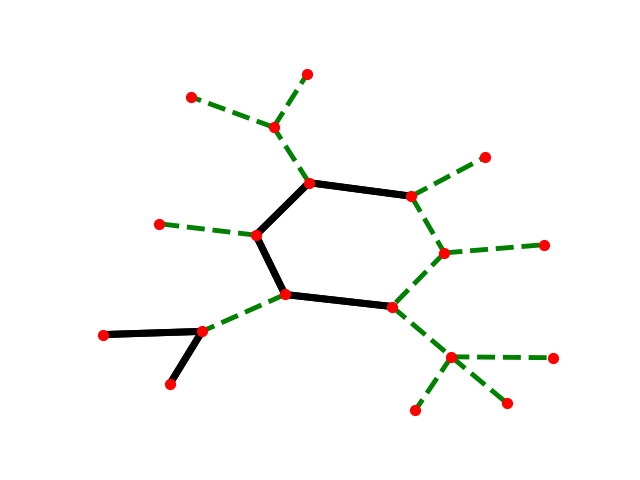} \\ \includegraphics[width=0.15\textwidth]{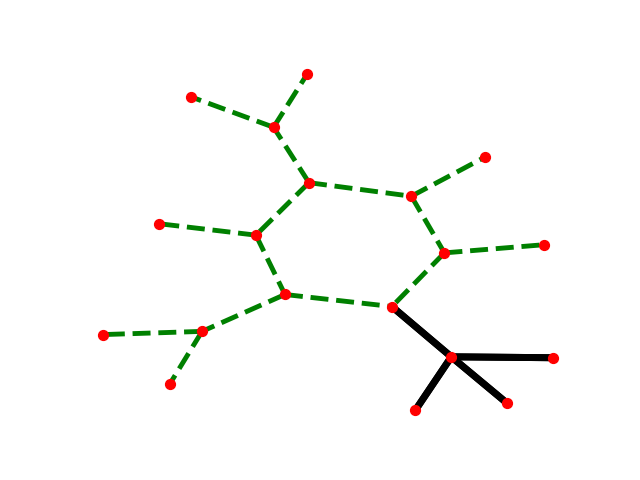}} \\
            \bottomrule
        \end{tabular}
        \caption{\small Examples of explanations produced with our method in the two datasets with different subgraph policies. DS has been used as Equivariant Feature Encoder.}
        \label{tab:examples}
    \end{table*}
    \egroup

    The results of our experiments are summarized in Table~\ref{tab:results}, where we also show the baseline results found using the original PGExplainer on a GIN. Even though we used the original implementation with the same settings, we obtained results that are different from those reported in the original paper~\cite{luo2020parameterized}. Some examples of explanations are shown in Table~\ref{tab:examples}. We notice that the explainer manages to correctly and tightly identify the relevant motifs in most scenarios, performing better or on par with the baselines. Some deeper insights follow.
    
    The explainer generally performs better on BA-2 than on Mutagenicity, both in terms of AUC and mask size (the expected mask size in BA-2 is around $20\%$). This is expected, given the simplicity and the synthetic nature of the dataset.
    
    In both datasets and both versions of EFE, the worst-performing subgraph policy is the one using ego networks (EN), with a very low average mask size. In our experiments, like in the original ESAN paper, we used a depth of $k=2$. In BA-2 the produced masks are too small to fully cover the ground truths, and the explainer struggles to uncover completely the target motif. In Mutagenicity on the other hand motifs are small, but graphs are quite large.
    When the masks $\M_i$ are too many, their individual contribution is washed out once they are combined. This might explain why the explanation $\mathcal{G}_{ex}$ struggles to highlight all the actual motifs (see figures in Table~\ref{tab:examples}).
    
    The EN+ policy performs slightly better than the standard EN, but there is a strong dependence on the architecture of EFE: it prefers the DS variant. This suggests that the improvement may be due to ESAN's performances rather than the explainer.
    
    Unlike what we see for EN and EN+, edge and node deletion policies on Mutagenicity often yield a higher average mask size, especially on larger graphs. The reason might be that with ED and ND, as the input gets larger, each edge appears more often in $\M_i$, even if it is not part of any~\mbox{motif}.

\section{Conclusion}
    We extended PGExplainer to ESAN, a recently proposed SGNN framework that leverages subgraphs to improve the expressiveness of GNNs. We showed that it is possible to adapt PGExplainer, so that it handles subgraphs and, with the right aggregation method, it is able to produce meaningful explanations for the whole input graph.

    As in the original work, PGExplainer still relies on what the model to be explained manages to learn: it is able to produce trustworthy explanations if ESAN identifies meaningful relations between nodes. In other words, the explainer fails if the underlying classifier solely relies on node features to predict the output. More research on interpretability for GNNs is needed in this regard.
    
    In our implementation, due to the structure of ESAN itself, the explainer only works on a partial embedding of the input graphs. In particular, the explainer does not make use of what $f_\text{SRL}$ and $f_\text{SE}$ learned. In future work, it could be interesting to take full advantage of all the embeddings produced by ESAN to overcome the difficulties we encountered when dealing with large graphs or small subgraphs. For example, the combination of all the masks into a single explanation could be conditioned on the final graph embedding.

\paragraph{Acknowledgments}
    The authors gratefully acknowledge NVIDIA Corporation for the donation of two RTX A6000 that were used in this project.

\bibliographystyle{abbrvnat}
\bibliography{references}

\end{document}